\documentclass[conference]{IEEEtran}
\usepackage{times}
\usepackage{xcolor}

\usepackage[numbers]{natbib}
\usepackage{multicol}
\usepackage[bookmarks=true]{hyperref}
\usepackage{graphicx}
\usepackage{algorithm}
\usepackage{algorithmic}
\usepackage{flafter}
\usepackage{amsmath,amssymb}
\usepackage[font=small]{caption}
\usepackage{booktabs}
\usepackage{tabularx}
\usepackage{array}

\newif\ifshowtodos
\showtodosfalse 

\newcommand{\blfootnote}[1]{%
  \begingroup
  \renewcommand{\thefootnote}{}%
  \footnotetext{#1}%
  \endgroup
}

\begin{document}


\title{It's Not Just More Demos: Counterfactual Action Sensitivity Coverage for Data-Efficient Robust Robot Imitation}


\author{Giovanni D’urso$^{1}$, Kaushik Roy$^{1}$, Nicholas Lawrance$^{1}$, Brendan Tidd$^{1}$}


%


\makeatletter
\let\@oldmaketitle\@maketitle 
\renewcommand{\@maketitle}{\@oldmaketitle 
  \begin{center}
    \includegraphics[width=0.95\textwidth]{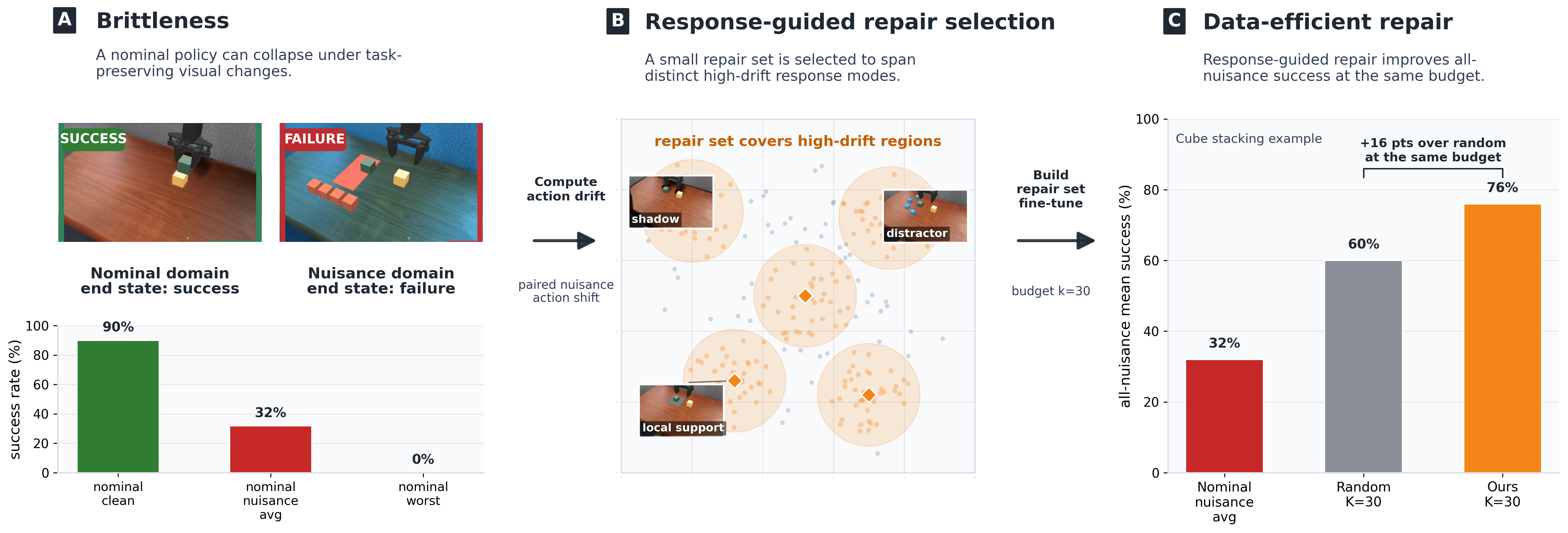}
    \captionof{figure}{\textbf{The Counterfactual Nuisance Behaviour Cloning (CFNBC) Pipeline.} 
    \textbf{(A)} A nominal visuomotor policy can succeed in the clean domain but fail under task-preserving visual nuisances. We measure \emph{action drift}: the change in the policy's predicted action between paired clean and nuisance observations. \textbf{(B)} Response-guided repair selection chooses a compact repair set that covers diverse high-drift response modes, rather than selecting redundant high-drift examples. \textbf{(C)} Fine-tuning on the selected repair set improves all-nuisance success more than random repair at the same budget. }
    \label{fig:hero_pipeline}
  \end{center}\vspace{-0.5em} 
}
\makeatother

\maketitle
\setcounter{figure}{1}

\begin{abstract}
\label{Sec:Abstract}
Visuomotor imitation learning has demonstrated success for manipulation tasks. However, the trained policies remain brittle to visual `nuisances', with even minor task-preserving variations such as lighting, distractions or changes in colour result in heavy degradation of the trained policy's performance. 
While increasing data diversity can improve robustness, it is unclear which additional demonstrations are informative for a particular trained policy. 
We propose Counterfactual Nuisance Behaviour Cloning (CFNBC), an offline data-selection framework for targeted robustness repair. Starting from a nominal policy trained on `clean' demonstrations, CFNBC generates paired clean and nuisance observations that preserve the expert action, then measures \emph{action drift}: the change in the policy's predicted action under a nuisance that should not alter the desired behaviour. This provides a policy-specific sensitivity signal for selecting a compact, response-diverse repair set from a larger candidate pool, without requiring rollout success labels or online policy execution.
We show in MuJoCo bimanual cube transfer and SimplerEnv cube stacking that action drift correlates with nuisance-induced failure, and that response-guided repair with only $20$--$30$ selected candidates substantially outperforms matched-budget random selection while approaching the performance of much larger random repair budgets.
These results support a data-centric view of robustness repair: the most useful data are not necessarily the most numerous, visually diverse, or obviously difficult, but the examples that cover fragile response modes of the current policy.

\blfootnote{$^1$ CSIRO Robotics, CSIRO, Australia. 
E-mails: {\tt\footnotesize \emph{firstname.lastname}@csiro.au}}
\end{abstract}

\IEEEpeerreviewmaketitle

\section{Introduction}
\label{Sec:Introduction}
Imitation learning approaches have become a practical route for training visuomotor manipulation policies, especially when designing rewards for online reinforcement learning is costly~\cite{osa2018algorithmic}. Recent bottlenecks have not been due to compute or policy architecture, but instead the quantity and quality of the data used for training and fine-tuning. Driven by this, the robotics community has pushed towards ever-larger data collection efforts, such as ALOHA Unleashed~\cite{zhao2024alohaunleashed} and the Open X-Embodiment dataset~\cite{o2024openxdataset}, which demonstrate the value of scaling robot data to develop robust or general policies. However, for many manipulation tasks, the more immediate question is: given a policy that already succeeds in a nominal setting, what additional data is needed to make it robust?

Unfortunately, real-world deployments are rarely ideal or nominal; thus, we study this question in the setting of behaviour cloning~(\emph{BC}), a standard imitation-learning objective that trains a policy to predict expert actions from demonstrated observations such as ACT~\cite{zhao2023learning}. These policies can perform well on clean demonstrations but fail under task-preserving visual nuisances. These nuisances, such as lighting changes, background appearance shifts, distractor objects or changed surface textures, can drastically alter the image while leaving the task state and desired expert action unchanged.  
A robust policy should be able to ignore such changes, yet low-data visuomotor policies respond to them in the same way as changes to the task itself. The typical approach is to collect data online or generate data through a domain randomisation process. However, random diversity can be inefficient because the examples may cover variations the policy already handles while missing specific visual conditions that induce brittle policy responses.

This inefficiency motivates a data-centric repair question: rather than blindly collecting more demonstrations, can we identify which nuisance conditions expose the missing data for a specific trained policy? We introduce Counterfactual Nuisance Behaviour Cloning (CFNBC), an offline repair data-selection framework for visuomotor imitation policies. Starting from a nominal policy and its clean demonstrations, CFNBC generates paired clean and nuisance observations that preserve the underlying task state and the expert action. We then measure the policy's \emph{action drift}: the change in the policy's predicted action between the clean and nuisance observations, computed in a normalised action space. Because the expert action should remain invariant, high action drift serves as a policy-specific sensitivity signal for task-irrelevant visual changes.

CFNBC uses this signal to select a compact repair set from a larger candidate response set. Instead of selecting only the highest-drift examples, which may be redundant, we select examples that cover diverse action-response modes under a fixed repair budget. The selected counterfactual examples are then paired with the original demonstration actions and used to fine-tune the nominal policy. In this way, counterfactual generation is not used as unstructured augmentation but as an offline audit to decide which repair examples are worth adding.

We evaluate CFNBC on two simulated manipulation tasks: MuJoCo bimanual cube transfer and SimplerEnv cube stacking. Across these tasks, we show that nominally successful ACT policies are brittle under task-preserving visual nuisances, that action drift strongly ranks fragile nuisance conditions, and that response-guided repair is substantially more data-efficient than random repair. In the low-budget regime, selected repair sets with $K=20$--$30$ candidates outperform matched-budget random selection and, in some cases, approach the performance of much larger random repair budgets.

In summary, our contributions are:
\begin{itemize}
    \item We formulate task-preserving visual nuisance repair as a data-selection problem for already-trained visuomotor imitation policies.
    \item We introduce action drift, an offline policy sensitivity signal computed from paired counterfactual observations without rollout success labels.
    \item We propose response-guided repair selection, which chooses compact repair sets that cover diverse high-sensitivity policy responses.
    \item We demonstrate on two manipulation tasks that response-guided repair substantially improves low-budget robustness compared with random repair data.
\end{itemize}

\section{Related Work}
\label{Sec:RelatedWork}
A common response to task-preserving visual shifts is to learn representations that collapse nuisance variation. Pretrained invariant representations and contrastive objectives such as InfoNCE~\cite{oord2018representationinfonce}, Time-Contrastive Networks~\cite{sermanet2018time}, R3M~\cite{nair2022r3m}, and Barlow Twins~\cite{zbontar2021barlow} can encourage invariance to some types of visual nuisances. However, in visuomotor manipulation, the split between nuisance and task-relevant visual information is task and state-dependent. The same visual information may be irrelevant in one state but crucial in another, such as shadows or local surface textures that provide information about object pose. If the assumed invariance is wrong, representation learning may remove information needed for control or retain nuisance correlations that still negatively affect the policy's actions~\cite{rabinovitz2021contrastiveDR,de2019causal}. 
Our work instead treats invariance as a policy-specific question. It uses task-preserving counterfactuals and changes in predicted actions to identify which visual changes actually perturb a trained policy's actions.

Domain randomisation and synthetic augmentation improve robustness by expanding the training distribution across a range of visual variations~\cite {tobin2017domain, garcia2023robust}. More targeted variants, such as active or automatic domain randomisation, bias sampling towards harder or more informative regions of the parameter space~\cite{mehta2020activedr,akkaya2019solvingautodr}. Counterfactual data augmentation methods such as RoCoDA~\cite{ameperosa2025rocoda} further exploit action-preserving transformations to generate synthetic demonstrations. These methods provide useful mechanisms for producing candidate repair data. However, they typically select or generate data at the level of environment parameters or visual transformations, rather than at the level of the policy's induced action response. CFNBC is complementary because it uses counterfactual generation to construct a candidate pool, then audits the trained policy to identify task-preserving visual changes that perturb its actions and selects repair examples that cover diverse, high-sensitivity response modes.

Active imitation-learning methods provide a policy-specific way to target failures. Methods such as DAgger~\cite{ross2011reductiondagger} execute the current policy and query an expert for corrective labels on states induced by that policy. Safety- or budget-aware variants reduce expert burden by requesting intervention only in risky, novel, or uncertain states~\cite{kelly2019hgdagger,hoque2021thriftydagger}. These methods directly address the problem of collecting data where the current policy is weak, but require online policy rollouts and, in many settings, human-in-the-loop correction. For robot hardware, this can be slow, expensive, and unsafe when the policy fails catastrophically. CFNBC targets the same policy-specific repair problem, but replaces online failure rollouts with an offline counterfactual action-sensitivity audit.

A complementary line of work reduces the cost of real-world evaluation by probing robot policies in simulation or real-to-sim benchmarks. Frameworks such as SimplerEnv~\cite{li2024evaluatingsimplerenv}, REALM~\cite{sedlacek2026realm}, GenGap~\cite{xie2024decomposinggengap}, and The Colosseum~\cite{pumacay2024colosseum} use controlled perturbations to study how robot policies degrade under visual, physical, or semantic distribution shifts before deployment.
These benchmarks can reveal a policy's fragility, but due to a persistent sim-to-real gap, naively generating simulated data and hoping to repair a real-world policy without degradation is difficult.
Our work uses simulation in a more diagnostic role, the paired counterfactual scenes estimate which nuisance variations change the policy's action, and response-space coverage is then used to prioritise real or synthetic repair examples under a limited data budget.

\section{Counterfactual Nuisance Behaviour Cloning}
\label{Sec:MethodCFNBC}
We propose Counterfactual Nuisance Behaviour Cloning~(CFNBC), an offline repair data-selection framework for visuomotor policies that succeed in a nominal domain but remain fragile under task-preserving visual nuisances. Given a clean demonstration set $\mathcal{D}_{clean}$ and a trained nominal policy $\pi_{\theta_0}$, CFNBC asks which counterfactual examples are most useful for repairing that specific policy under a limited data budget.

CFNBC proceeds in three stages. First, we generate paired clean and nuisance observations for the same task state (preserving the expert action), and measure the policy's \emph{action drift} and response between each pair. The action drift and response features are used to define a finite candidate response set, where each candidate is a sampled counterfactual response window from a demonstration trajectory rather than an entire new demonstration and is described by a drift score and a policy response feature. Second, we select a compact repair set that covers diverse high-sensitivity response modes, rather than simply choosing the largest drift examples. Finally, we pair the selected nuisance observations with their inherited (expert) demonstration actions and fine-tune the nominal policy on the resulting counterfactual repair data.

\subsection{Task-preserving counterfactuals}

At timestep $t$, the policy receives input $s_t=(o_t,q_t)$, where $o_t$ is an RGB observation and $q_t$ is the robot state. Let $x_t$ denote the underlying task state, such as object pose and robot pose, and let $\eta$ denote visual nuisance factors such as lighting, distractors, or surface appearance. We write the observation as $o_t=g(x_t,\eta)$. For a clean visual setting $\eta_c$ and nuisance setting $\eta_n$, the paired observations are
\[
    s_t^c=(g(x_t,\eta_c),q_t),
    \qquad
    s_t^n=(g(x_t,\eta_n),q_t).
\]
A nuisance is task-preserving when it changes the visual observation but leaves the desired expert action unchanged:
\begin{equation}
    a^*(s_t^c)=a^*(s_t^n).
    \label{eq:nuisance_action_equivalence}
\end{equation}
This paired construction isolates visual sensitivity: if the task state and expert action are unchanged, then a large change in the policy's predicted action indicates sensitivity to a task-irrelevant visual shift.

\subsection{Action drift and candidate response set}
For each sampled candidate response \(c_i\), indexed by timesteps $t\in\mathcal{T}_i$, we generate paired clean and nuisance inputs. Querying the nominal policy gives
\[
    a^c_{i,t}=\pi_{\theta_0}(s^c_{i,t}),
    \qquad
    a^n_{i,t}=\pi_{\theta_0}(s^n_{i,t}).
\]
Because different action dimensions may have different scales, we compare actions in a normalised action space. Let $N(\cdot)$ denote per-dimension normalisation using action statistics from the training demonstrations. The normalised action delta and scalar action drift are
\begin{equation}
    \Delta a_{i,t}=N(a^n_{i,t})-N(a^c_{i,t}),
    \qquad
    d_i=\frac{1}{|\mathcal{T}_i|}
    \sum_{t\in\mathcal{T}_i}\|\Delta a_{i,t}\|_2 .
    \label{eq:action_drift}
\end{equation}
The drift score $d_i$ is not an expert action error: both predictions may differ from the demonstrated action. Instead, it measures how much the learned policy changes its action in response to a task-preserving visual nuisance. A large drift indicates that the policy is sensitive to visual changes even without rollout success labels.

We also construct a response feature $r_i$ from the sequence $\{\Delta a_{i,t}\}_{t\in\mathcal{T}_i}$, using summary statistics across timesteps and action dimensions. The scalar $d_i$ measures sensitivity, while $r_i$ is used to compare candidates by the type of policy response they induce. Sampling demonstrations, timesteps, and nuisance settings produce a finite candidate response set $\mathcal{C}$. Each candidate $c_i\in\mathcal{C}$ stores the paired nuisance inputs, inherited expert actions, action drift score $d_i$, and response feature $r_i$.

\subsection{Response-guided repair selection}
We assume a limited repair budget $K$, reflecting the cost of generating, selecting, or collecting additional repair data. The goal is therefore not to train on every possible nuisance, but to identify which counterfactual examples are most useful for repairing the current policy.
A simple repair strategy is to select the $K$ candidates with the largest action drift. However, high-drift examples may be redundant if they correspond to the same failure mode. CFNBC instead selects a compact repair set $\mathcal{R}_K\subseteq\mathcal{C}$ that covers diverse policy response modes under a fixed budget.

Let $A(r_i,r_j)$ be a response affinity function that is large when candidates $i$ and $j$ induce similar policy responses. The budget-feasible family is
\begin{equation}
    \mathcal{B}_K=\{\mathcal{R}\subseteq\mathcal{C}:|\mathcal{R}|\le K\}.
\end{equation}
We select the repair set by maximising weighted response coverage:
\begin{equation}
    \mathcal{R}_K
    =
    \operatorname*{arg\,max}_{\mathcal{R}\in\mathcal{B}_K}
    \sum_{i\in\mathcal{C}}
    w_i
    \max_{j\in\mathcal{R}} A(r_i,r_j).
    \label{eq:repair_set_selection}
\end{equation}
The weights $w_i$ can be uniform, which favours broad response diversity, or can increase with $d_i$, which prioritises high-sensitivity response modes. In practice, we greedily optimise Eq.~\ref{eq:repair_set_selection} by sequentially adding the candidate with the largest marginal improvement until $K$ candidates have been selected. 
We use temporally subsampled candidate windows rather than all frames in the dataset; implementation details of the response feature, affinity function, and drift weighting are provided in Appendix~\ref{Apx:Algorithm}.


\subsection{Counterfactual-Based Repair Fine-Tuning}
\label{SubSec:MethodFineTuning}
Once the repair set $\mathcal{R}_K$ has been selected, each selected candidate contributes nuisance observations paired with the original expert actions from the clean demonstrations. For each $c_i \in \mathcal{R}_K$ and timestep $t\in\mathcal{T}_i$, let $s^n_{i,t}$ denote the nuisance input and $a_{i,t}$ denote the corresponding clean demonstration action. Because the nuisance is task-preserving, $a_{i,t}$ remains a valid target for the nuisance input. The selected counterfactual repair dataset is
\begin{equation}
    \mathcal{D}^{cf}_{\mathcal{R}_K}
    =
    \{(s^n_{i,t}, a_{i,t}) :
    c_i \in \mathcal{R}_K,\;
    t \in \mathcal{T}_i\}.
    \label{eq:counterfactual_repair_dataset}
\end{equation}

Starting from the nominal policy parameters $\theta_0$, we fine-tune on a mixture of the original clean demonstrations and selected counterfactual repair examples:
\begin{equation}
    \theta_{\mathrm{repair}}
    =
    \arg\min_{\theta} \left[
    \mathcal{L}_{BC}(\theta;\mathcal{D}_{clean})
    +
    \lambda_{cf}
    \mathcal{L}_{BC}(\theta;\mathcal{D}^{cf}_{\mathcal{R}_K})\right].
    \label{eq:cfnbc_objective}
\end{equation}

The repaired policy is \(\pi_{\theta_{\mathrm{repair}}}\). This fine-tuning step uses no new expert labels, rollout success labels, or rewards: the repair targets are inherited from the original demonstrations. At the same time, the selection signal arises from the nominal policy's own changes in action in response to paired counterfactual inputs. Thus, CFNBC separates the offline identification of fragile response modes from the cost of collecting or training on large amounts of unstructured repair data.

\section{Experiment Setup}
\label{Sec:ExperimentSetup}
To evaluate the efficacy of Counterfactual Nuisance Behaviour Cloning~(CFNBC), we systematically assess how well various data-selection strategies recover policy robustness to targeted, task-preserving visual distribution shifts while adhering to data budgets.

\noindent\textbf{Tasks and nominal policies.}
We evaluate on two simulated manipulation tasks using Action Chunking with Transformers (ACT)~\cite{zhao2023learning} as the base policy architecture. \textbf{Bimanual cube transfer} involves two simulated arms in MuJoCo~\cite{todorov2012MuJoCo}. One arm picks up the cube and hands it to the other using 14-dimensional joint-space actions. We train the nominal policy from 50 clean demonstrations, denoted A50. In \textbf{SimplerEnv cube stacking}~\cite{li2024evaluatingsimplerenv}, a single WidowX arm simulated with SAPIEN~\cite{sapien2020} must stack a green cube on a yellow cube using 7-dimensional delta end-effector actions $(\texttt{XYZRPYG})$. We train the nominal policy from 200 clean demonstrations, denoted A200. Both nominal policies achieve approximately $90\%$ clean-domain success before nuisance evaluation.

\noindent\textbf{Task-preserving nuisance space.} 
We evaluate robustness over a structured nuisance space containing four task-preserving visual factors: appearance changes to the table and background, lighting and shadow changes, irrelevant distractor objects near the target, and local support changes such as a cloth patch under the manipulated objects. Each factor has four severity levels ($L0-L3$), giving 16 single-factor conditions, and we additionally evaluate six predefined multi-factor combination conditions, for 22 nuisance conditions in total. These conditions alter the visual observation while preserving the underlying task state and the validity of the demonstrated action. The levels are ordered within each factor by intended severity, but are not calibrated across factors; combination conditions are treated as predefined stress tests. Representative examples are shown in Figure~\ref{fig:nuisance_examples}, with full nuisance definitions in Appendix~\ref{Apx:Nuisance}.

\begin{figure*}[t]
    \centering
    \includegraphics[width=0.95\textwidth]{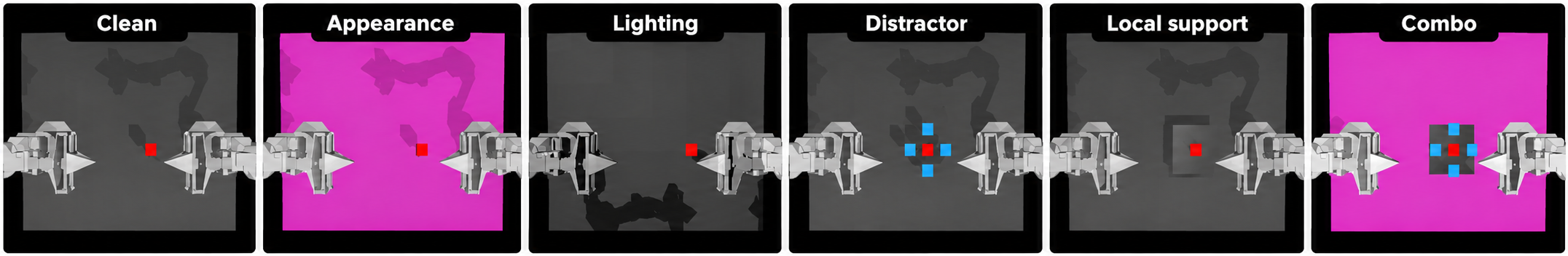}
    \includegraphics[width=0.95\textwidth]{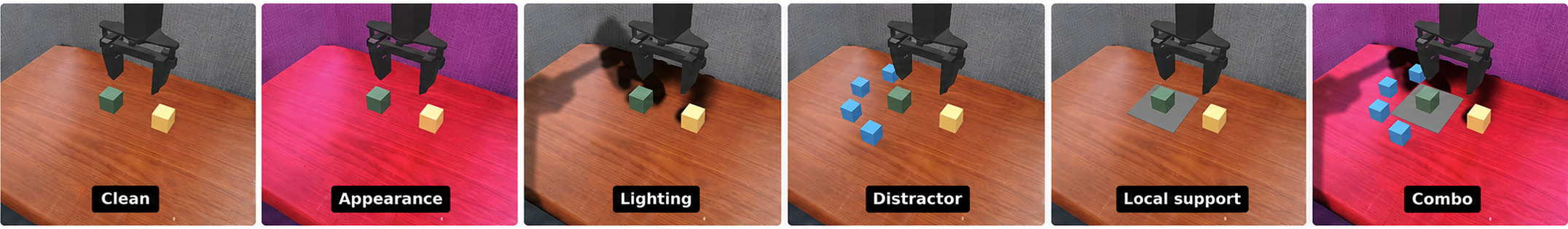}
    \caption{
    Representative task-preserving nuisance conditions. We show one example task for visual clarity: clean, appearance $L3$, lighting $L3$, distractor $L3$, local support $L3$, and one predefined multi-factor combination. The same nuisance families are instantiated across both evaluation tasks; full nuisance grids are provided in Appendix~\ref{Apx:Nuisance}.
    }
    \label{fig:nuisance_examples}
\end{figure*}

\begin{table*}[th]
\centering
\caption{Nominal policies are brittle under task-preserving visual nuisances. All values are success rates in $[0,1]$. Both policies achieve high in-domain success, but performance drops sharply under nuisance shifts that preserve the intended expert action.}
\label{tab:nominal_brittleness}
\begin{tabular}{llrrrrrrrrrr}
\toprule
Task & Policy/data & Clean & App. & Light & Distr. & Support & Combo & Nuis. mean & Worst & Gap mean & Gap worst \\
\midrule
Cube transfer & ACT / A50 & 0.96 & 0.09 & 0.44 & 0.33 & 0.22 & 0.39 & 0.30 & 0.04 & 0.66 & 0.92 \\
Cube stack & ACT / A200 & 0.90 & 0.71 & 0.57 & 0.00 & 0.41 & 0.03 & 0.32 & 0.00 & 0.58 & 0.90 \\
\bottomrule
\end{tabular}

\vspace{0.4em}
\begin{minipage}{0.98\linewidth}
\footnotesize
App., Light, Distr., and Support are means over the four levels of each nuisance factor. Combo is the mean over the six predefined multi-factor conditions. Nuis. mean averages all 22 nuisance conditions. Worst is the minimum success over the 22 conditions. Gaps are the drop in performance relative to Clean.
\end{minipage}
\end{table*}

\noindent\textbf{Candidate response sets and repair selection.}
For each nominal policy, we construct a candidate response set $\mathcal{C}$ by applying task-preserving nuisance variations to nominal demonstration episodes and querying $\pi_{\theta_0}$ on paired clean and nuisance inputs. Each candidate is assigned an action drift score and response feature as described in Section~\ref{Sec:MethodCFNBC}. The main comparison evaluates response-guided repair against matched-budget random repair and top-drift selection, which selects the highest-drift candidates without explicitly encouraging response diversity. 

\noindent\textbf{Fine-tuning and evaluation.}
For each selected repair set, we generate the corresponding counterfactual repair data and fine-tune the nominal policy from the same checkpoint using the same training procedure. Thus, differences in downstream performance reflect the selected repair data rather than changes in optimisation or initialisation. We evaluate each repaired policy by rollout success over 100 new episodes for the clean condition and each of the 22 nuisance conditions. We also evaluate held-out nuisance conditions with different colours, shadow angles, distractor shapes, and distractor positions; these are not used during candidate response generation, repair selection, or fine-tuning. Candidate-pool sizes and held-out definitions are reported in Appendix~\ref{Apx:Nuisance}.

\noindent\textbf{Training seeds and uncertainty.}
The main repair results are averaged over three initialisation seeds. We use these means to reduce dependence on a single run, but do not claim statistical significance with $n=3$; larger-scale evaluation should assess sensitivity to the training seed, random repair, the candidate pool, and hyperparameters.

\section{Results and Discussion}
\label{Sec:Results}
In this section, we present our simulation results to evaluate the efficacy of CFNBC. We address whether action drift is a reliable proxy for fragility, how efficiently we can cover the response space, and the resulting data efficiency of targeted repair compared to random data collection.
Together these experiments test the central claim that useful examples that expose diverse high-sensitivity responses result in more robust policies than simply more uninformed examples.

\subsection{Nominal policies are brittle under task-preserving nuisances}
We first present the nominal policies $\pi_{\theta_0}$ before repair. Both policies achieve high success rates on their training distributions, but their performance degrades heavily when task-preserving nuisances are introduced. Since these nuisances preserve the intended expert action, a robust policy should preserve the intended action and achieve a similar success rate across the clean and nuisance conditions. The observed performance gaps motivate targeted repair, despite nuisance shifts that preserve the intended action.

\begin{figure}[t]
    \centering
    \includegraphics[width=0.9\linewidth]{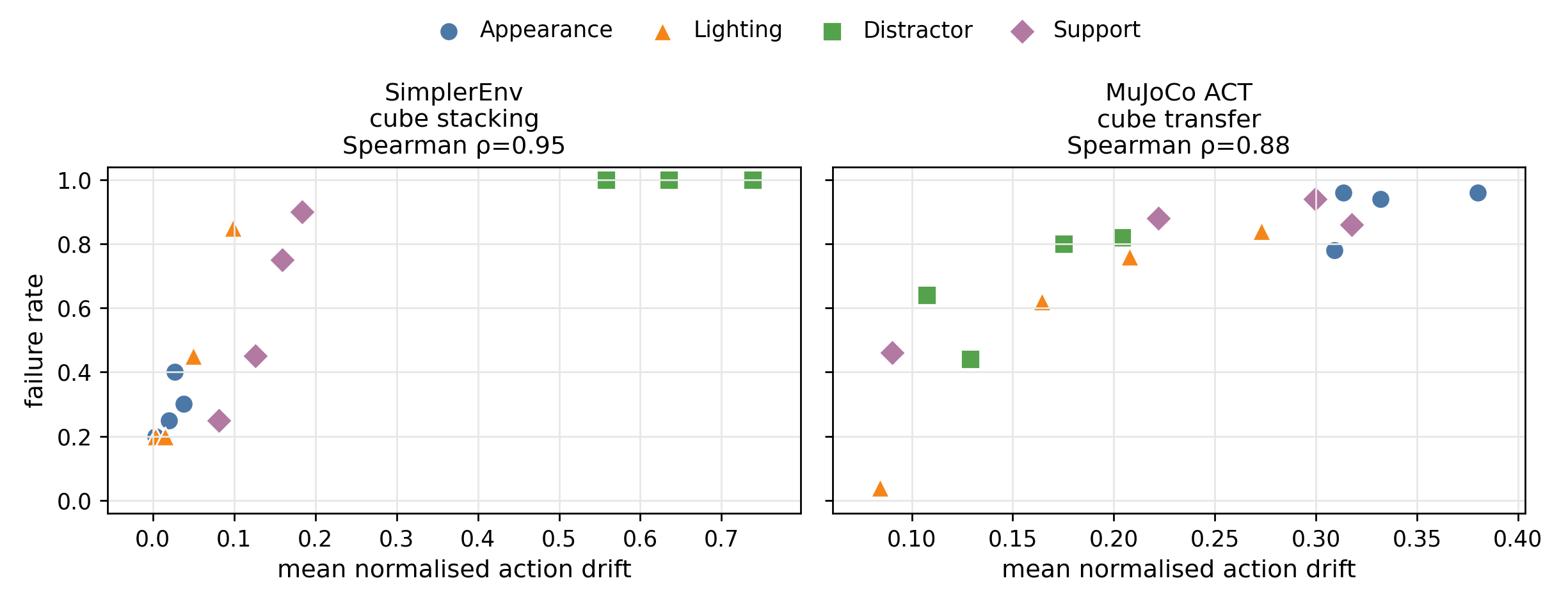}
    \caption{
Action drift identifies fragile nuisance conditions. Each point is a nuisance condition; the x-axis shows mean normalised action drift computed from paired candidates in the candidate response set, and the y-axis shows rollout failure rate under that condition. Spearman correlations are computed over the plotted conditions and show that action drift provides a strong offline ranking signal for policy fragility to the specific nuisance condition.
}
    \label{fig:action_drift_correlation}
\end{figure}

\subsection{Action drift provides an offline fragility signal}
To use our framework for offline data selection, we require a signal that ranks nuisance conditions by policy fragility without requiring environment rollouts. Given a nominal policy trained on nominal demonstrations, we compute the normalised action drift across paired counterfactual states. 

As shown in Figure~\ref {fig:action_drift_correlation}, nuisance conditions with higher normalised action drift tend to have higher rollout failure rates in the corresponding evaluation environment. 
This relationship holds across both the SimplerEnv cube stacking and MuJoCo cube transfer tasks, with Spearman rank correlations (a measure of order preservation) of $\rho=0.95$ and $\rho=0.88$, respectively, supporting the use of the signal to identify fragile nuisance conditions.
Crucially, performing this comparison in the normalised action space prevents the scalar magnitude of some action dimensions from dominating the metric, thereby providing a rollout-free offline signal of fragility under nuisance perturbations.

This ranking should be interpreted at the level of individual nuisance conditions rather than as a strict ordering of the semantic nuisance families. 
For example, a mild appearance shift can be more damaging than close proximity of distractors for one policy, but not so for another. Since the impact of a nuisance depends on the specific model and task, this supports the need for a policy-specific audit. In the cube stacking task, the distractor and local support conditions are the most damaging, but in the cube transfer task, appearance and local support are more damaging. Action drift, therefore, helps to identify fragile conditions without requiring rollout evaluation of the policy.

\begin{table*}[t]
\centering
\caption{
Response-guided repair improves robustness under limited repair budgets. Success columns report three-seed-averaged rates in $[0,1]$. Single-factor averages the single-factor nuisance conditions, while Combo averages the predefined multi-factor nuisance conditions. All nuis. averages all nuisance conditions excluding clean. Gain vs nominal is the improvement in All nuis. over the nominal policy for the same task.
}
\label{tab:data_efficiency_summary}
\begin{tabular}{llrrrrrrr}
\toprule
Task & Method & $K$ & Clean & Single-factor & Combo & All nuis. & Worst & Gain vs nominal \\
\midrule
Cube transfer & Nominal & 0 & 0.96 & 0.27 & 0.39 & 0.30 & 0.04 & 0.00 \\
Cube transfer & Random & 20 & 0.80 & 0.60 & 0.44 & 0.56 & 0.07 & 0.26 \\
Cube transfer & Top-drift & 20 & 0.90 & 0.76 & 0.69 & 0.74 & 0.20 & 0.44 \\
Cube transfer & Response-diverse (ours) & 20 & 0.97 & 0.94 & 0.99 & 0.96 & 0.53 & 0.66 \\
Cube transfer & Random (high-budget) & 500 & 1.00 & 1.00 & 1.00 & 1.00 & 1.00 & 0.70 \\
\midrule
Cube stacking & Nominal & 0 & 0.90 & 0.42 & 0.03 & 0.32 & 0.00 & 0.00 \\
Cube stacking & Random & 30 & 0.87 & 0.67 & 0.40 & 0.60 & 0.10 & 0.28 \\
Cube stacking & Factor-balanced random & 30 & 0.93 & 0.74 & 0.34 & 0.63 & 0.13 & 0.31 \\
Cube stacking & Top-drift & 30 & 0.92 & 0.72 & 0.37 & 0.63 & 0.22 & 0.31 \\
Cube stacking & Drift-weighted response (ours) & 30 & 0.90 & 0.84 & 0.54 & 0.76 & 0.20 & 0.44 \\
Cube stacking & Random (high-budget) & 500 & 0.88 & 0.88 & 0.91 & 0.89 & 0.75 & 0.57 \\
\bottomrule
\end{tabular}
\end{table*}

\subsection{Response-guided selection improves data efficiency}
We next compare response-guided repair selection against random and high-drift baselines under matched low-budget repair settings. For each repair budget $K$, the policy is fine-tuned using $K$ selected counterfactual repair candidates and evaluated across nuisance conditions. The key question is whether selecting examples by policy response can recover robustness using fewer repair examples than random selection.

\begin{figure}[ht]
    \centering
    \includegraphics[width=0.98\linewidth]{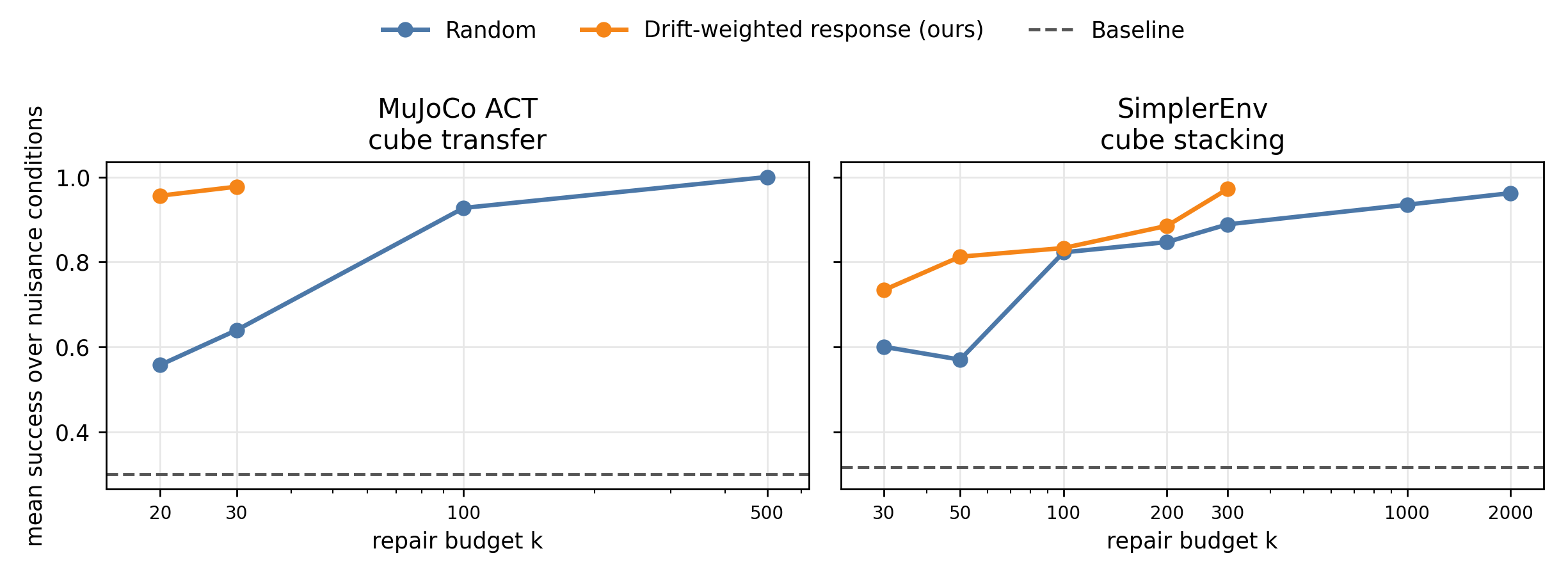}
    \caption{
    Response-guided repair improves data efficiency. Each point shows the three-seed-averaged mean success rate over nuisance conditions after fine-tuning with $K$ selected counterfactual repair candidates. Dashed horizontal lines show the nominal policy before repair. Response-guided selection reaches high robustness with substantially fewer repair candidates than random selection, while large random budgets can eventually close the gap.
    }
    \label{fig:data_efficiency}
\end{figure}

Figure~\ref{fig:data_efficiency} shows that response-guided selection is most valuable in the low-budget regime. In Cube transfer, our method nearly restores full performance in the tested nuisance domains with $20$--$30$ repair candidates, while random needs around $100$--$500$ samples to match this success rate. In cube stacking, the results improve more slowly with a large marginal benefit for the initial samples, but response-guided selection still reaches high nuisance success with fewer repair candidates than random selection. At much larger budgets, random also reaches the performance ceiling, indicating that data volume can compensate for uninformed selection when enough data is available. 

Table~\ref{tab:data_efficiency_summary} provides numerical results comparing our method against some baselines. At matched low budgets, response-guided repair improves the all nuisance success rate by $+0.66$ in Cube transfer and $+0.44$ in Cube stacking, compared with $+0.26$ and $+0.28$ for random selection. The repaired policies also preserve clean performance, suggesting that the selected counterfactual examples improve robustness to nuisances without simply trading away nominal behaviour.
Additionally, the gains on Combo and Worst further indicate that the repair is not limited to a single nuisance family, but improves broader robustness across the evaluated nuisance set.

Held-out transfer is partial: response-guided repair improves transfer in Cube transfer, while SimplerEnv methods are closer together, and large random budgets transfer best. This suggests that CFNBC selects useful examples from the available candidate response set, but still depends on candidate-set coverage of deployment nuisances. Full held-out results are reported in Appendix~\ref{Apx:HeldOut}.

\subsection{Practical Lessons and Limitations}
A core takeaway is that robust imitation learning depends not only on how much additional data is collected, but on whether that data covers the policy's missing response modes. Random repair data can eventually improve robustness, but may spend many samples on nuisance variations that the policy already handles. In contrast, action drift provides an offline audit of where a trained policy changes its behaviour under task-preserving visual shifts. This makes it possible to prioritise additional demonstrations or counterfactual repair examples that are likely to address specific policy weaknesses.

Although our experiments use selected counterfactual samples directly for fine-tuning,
CFNBC can also be interpreted as a practical data-acquisition policy. In a real-robot workflow, an operator could start from nominal demonstrations, generate or observe candidate nuisance variants, compute action drift offline using the current policy, and then prioritise high-value scenarios for real robot replay or additional teleoperated demonstration. We have not yet validated this collection loop on hardware, and an important open question is whether direct synthetic repair is sufficient or whether selected nuisance scenarios must be recreated with real robot demonstrations.

The main limitation is that CFNBC is only as good as the candidate response set and the task-preserving assumption. The nuisance interventions used in this work are manually constrained so that the task state, goal, robot state, and inherited expert action remain valid. We do not claim that all possible distractor or support changes are task-preserving; CFNBC requires candidate interventions to be constructed or filtered so that inherited action labels remain meaningful. If deployment contains a nuisance-induced response mode absent from $\mathcal{C}$, the selector cannot identify or repair it.

Action drift should also be interpreted as a sensitivity signal rather than a task-agnostic success predictor or safety guarantee. 
Real robot use would still require independent safety constraints, monitoring, and validation. Additionally, for multimodal tasks or policies, large drift may correspond to a valid alternative action rather than a failure. For stochastic or diffusion-style policies, drift should be estimated over multiple action samples or distributional summaries rather than a single prediction. Conversely, small local drift may miss delayed failures, hesitation, or closed-loop compounding errors, while large local drift may be recoverable by a robust closed-loop policy. Future work should therefore study richer candidate generation, hardware validation, larger-scale statistical evaluation, and combinations with other offline signals such as ensemble uncertainty, policy disagreement, or small amounts of rollout-derived feedback.

\section{Conclusion}
\label{Sec:Conclusion}

CFNBC reframes robust imitation learning as an offline data-selection problem: not only asking whether we have enough data, but whether the data we collect covers the response modes where the current policy is fragile. By measuring action drift under task-preserving counterfactuals, CFNBC identifies compact repair sets that improve robustness with far fewer examples than random repair data. This provides a practical step toward data-efficient robustness repair for visuomotor policies, and opens future work on hardware deployment, broader policy architectures, and online demonstration scoring.


\bibliographystyle{plainnat}
\bibliography{references}

\appendices
\label{Sec:Appendix}

\section{CFNBC Algorithm and Implementation Details} \label{Apx:Algorithm} 
This appendix provides the full CFNBC procedure used in the experiments, including candidate response set construction, response-guided repair selection, and counterfactual repair fine-tuning. The main paper presents the core equations; Algorithm~\ref{alg:cfnbc} provides the complete workflow for generating the selected repair set $\mathcal{R}_K$ and the repaired policy $\pi_{\theta_{\mathrm{repair}}}$.

\noindent\textbf{Candidate sampling.} For each task policy, we construct a candidate response set $\mathcal{C}$ by applying the nuisance conditions to nominal demonstration episodes and querying the nominal policy on paired clean and nuisance inputs. For the SimplerEnv task, we sampled a candidate pool of 2200 paired episodes and evaluated every 10th timestep, yielding $26,125$ samples.
For the MuJoCo bimanual cube transfer task, we sampled a candidate pool of 50 paired episodes to yield $44,000$ samples due to the longer demonstrations. Computing these sets requires approximately 15 minutes using an A2000 GPU.
 
\noindent\textbf{Response features.} Each candidate response feature $r_i$ summarises the sequence of normalised action deltas $\{\Delta a_{i,t}\}_{t\in\mathcal{T}_i}$. In our experiments, this feature contains summary statistics over the sampled timesteps and action dimensions, including mean, standard deviation, mean absolute value, and maximum absolute value. These features are used for comparing candidates by the type of policy response they induce, while $d_i$ stores the scalar magnitude of that response. 

\noindent\textbf{Greedy repair-set selection.} The response-guided selector is optimised greedily. Starting from $\mathcal{R}=\emptyset$, we repeatedly add the candidate with the largest marginal improvement in the weighted response-coverage objective until $|\mathcal{R}|=K$. This approximates the budgeted selection objective used in the main paper and encourages the selected repair set to cover diverse high-sensitivity response modes. 

\noindent\textbf{Drift-weighted response selection.}
For the reported response-guided selector, each candidate \(c_i\) is represented by a compact response feature \(r_i\) computed from the normalised action-delta sequence. We concatenate four per-action-dimension summary vectors over sampled timesteps: mean, standard deviation, mean absolute value, and maximum absolute value. Response features are standardised before computing affinities. Candidate affinities are computed with an RBF kernel,
\[
    A_{ij}
    =
    \exp\left(
    -\frac{\|\tilde r_i-\tilde r_j\|_2^2}{2\sigma^2}
    \right),
\]
where \(\tilde r_i\) denotes the standardised response feature and \(\sigma\) is set to the median non-zero pairwise distance in the candidate pool.

The scalar drift score is
\[
    d_i
    =
    \frac{1}{|\mathcal{T}_i|}
    \sum_{t\in\mathcal{T}_i}
    \|\Delta a_{i,t}\|_2 ,
\]
and we normalise it by the mean drift in the candidate pool, giving
\[
    \tilde d_i = \frac{d_i}{\bar d}.
\]
In the drift-weighted selector, \(\tilde d_j\) weights the importance of covering high-drift response modes, while \(\tilde d_i\) acts as a candidate quality multiplier. At each greedy step, with the current coverage
\[
    m_j = \max_{c_\ell \in \mathcal{R}} A_{\ell j},
\]
candidate \(c_i\) is scored by
\[
    g(i \mid \mathcal{R})
    =
    \tilde d_i
    \sum_j
    \tilde d_j
    \left[
        \max(m_j, A_{ij}) - m_j
    \right].
\]
The selected repair set, therefore, favours candidates that cover response patterns not already represented in \(\mathcal{R}\), with an additional bias toward nuisance examples that induce large changes in the nominal policy's predicted action.

\noindent\textbf{Fine-tuning protocol.} All repaired policies are fine-tuned from the same nominal checkpoint $\theta_0$. The selected counterfactual repair data are paired with inherited expert actions from the original clean demonstrations and mixed with $\mathcal{D}_{clean}$. This ensures that differences between repaired policies are attributable to the selected repair data rather than to changes in the optimisation procedure. 
We fine-tune the models for 800 epochs from their nominal trained states and use a $\lambda_{cf}=0.5$ for the reported experiments. In preliminary sweeps, values in approximately $0.5$--$0.75$ balanced nuisance repair with retention of clean-domain performance, while larger weights tended to degrade nominal behaviour, consistent with catastrophic forgetting. These hyperparameters are not optimised and could be investigated in future work.

\begin{algorithm}[t]
\caption{Counterfactual Nuisance Behaviour Cloning (CFNBC)}
\label{alg:cfnbc}
\textbf{Input:} Clean demonstrations $\mathcal{D}_{clean}$, nominal policy $\pi_{\theta_0}$, nuisance generator $G$, action normalisation $N$, repair budget $K$, counterfactual weight $\lambda_{cf}$ \\
\textbf{Output:} Selected repair set $\mathcal{R}_K$, repaired policy $\pi_{\theta_{\mathrm{repair}}}$
\begin{algorithmic}[1]
\STATE Initialise candidate response set $\mathcal{C} \leftarrow \emptyset$
\FOR{each sampled demonstration segment $i$ with timesteps $\mathcal{T}_i$}
    \STATE Sample a task-preserving nuisance condition $\eta_i$
    \FOR{each timestep $t \in \mathcal{T}_i$}
        \STATE Generate paired nuisance input $s^n_{i,t} \leftarrow G(s^c_{i,t}, \eta_i)$
        \STATE Query nominal policy:
        $a^c_{i,t} \leftarrow \pi_{\theta_0}(s^c_{i,t})$,
        $a^n_{i,t} \leftarrow \pi_{\theta_0}(s^n_{i,t})$
        \STATE Compute action delta:
        $\Delta a_{i,t} \leftarrow N(a^n_{i,t}) - N(a^c_{i,t})$
    \ENDFOR
    \STATE Compute drift score $d_i$ from $\{\Delta a_{i,t}\}_{t\in\mathcal{T}_i}$
    \STATE Compute response feature $r_i$ from $\{\Delta a_{i,t}\}_{t\in\mathcal{T}_i}$
    \STATE Add candidate $c_i$ to $\mathcal{C}$
\ENDFOR
\STATE Compute response similarity $A$ from response features $\{r_i\}$
\STATE Select repair set $\mathcal{R}_K \subseteq \mathcal{C}$ by greedily maximising Eq.~\ref{eq:repair_set_selection}
\STATE Construct selected counterfactual repair dataset:\\
$
    \mathcal{D}^{cf}_{\mathcal{R}_K}
    =
    \{(s^n_{i,t}, a_{i,t}) :
    c_i \in \mathcal{R}_K,\;
    t \in \mathcal{T}_i\}
$
\STATE Fine-tune from $\theta_0$:\\
$
    \theta_{\mathrm{repair}}
    \leftarrow
    \arg\min_\theta
    \left[\mathcal{L}_{BC}(\theta;\mathcal{D}_{clean})
    +
    \lambda_{cf}
    \mathcal{L}_{BC}(\theta;\mathcal{D}^{cf}_{\mathcal{R}_K})\right]
$
\RETURN $\mathcal{R}_K, \pi_{\theta_{\mathrm{repair}}}$
\end{algorithmic}
\end{algorithm}

\section{Task-Preserving Nuisance Definitions}
\label{Apx:Nuisance}

This appendix documents the nuisance factors used to construct paired clean and counterfactual observations. All nuisance conditions are designed to alter the visual observation while preserving the underlying task state, success condition, and validity of the demonstrated expert action. The nuisance levels $L0$--$L3$ are ordered within each factor by intended visual severity, but are not calibrated across factors; for example, $L3$ lighting is not assumed to be equally difficult as $L3$ distractor proximity.

\begin{table*}[t]
\centering
\caption{
Summary of task-preserving nuisance factors. Each factor changes the visual observation while preserving the underlying task state and the expert action label inherited from the clean demonstration.
}
\label{tab:nuisance-summary}
\small
\renewcommand{\arraystretch}{1.15}
\begin{tabularx}{\textwidth}{
    >{\raggedright\arraybackslash}p{0.16\textwidth}
    >{\centering\arraybackslash}p{0.10\textwidth}
    >{\raggedright\arraybackslash}X
    >{\raggedright\arraybackslash}X
}
\toprule
Nuisance family & Levels & Perturbation & Task-preserving assumption \\
\midrule
Appearance
& $L0$--$L3$
& Changes to table, background, or scene colours/textures.
& Object poses, robot state, goal, and success condition are unchanged. \\

Lighting and shadow
& $L0$--$L3$
& Changes to light direction and resulting shadow direction or intensity.
& Scene geometry and required action are unchanged. \\

Distractor proximity
& $L0$--$L3$
& Irrelevant blue distractor objects placed at varying distances from the target.
& Distractors are non-target objects and do not change the demonstrated task objective. \\

Local support
& $L0$--$L3$
& Local surface/support changes near the manipulated objects, such as a grey cloth patch.
& The support appearance changes without changing the intended manipulation action. \\

Combination
& six settings
& Multi-factor stress tests combining several nuisance families.
& Combined perturbations preserve the underlying task state and expert action. \\
\bottomrule
\end{tabularx}
\end{table*}

The 16 single-factor nuisance conditions are formed by applying the four nuisance families at four levels each. We additionally evaluate six predefined multi-factor combination conditions, giving 22 nuisance conditions in total. The combination conditions are used as stress tests rather than as an exhaustive factorial decomposition of all possible nuisance interactions. Specifically, the six combination conditions are: \begin{itemize} \item appearance $L3$ + local support $L3$; \item distractor proximity $L3$ + local support $L3$; \item appearance $L2$ + local support $L2$; \item appearance $L3$ + distractor proximity $L3$ + lighting/shadow $L3$ + local support $L3$; \item appearance $L2$ + distractor proximity $L2$ + lighting/shadow $L2$ + local support $L2$; \item appearance $L3$ + distractor proximity $L2$ + lighting/shadow $L3$ + local support $L2$. \end{itemize}

\begin{figure*}[p]
    \centering
    \includegraphics[width=0.98\textwidth]{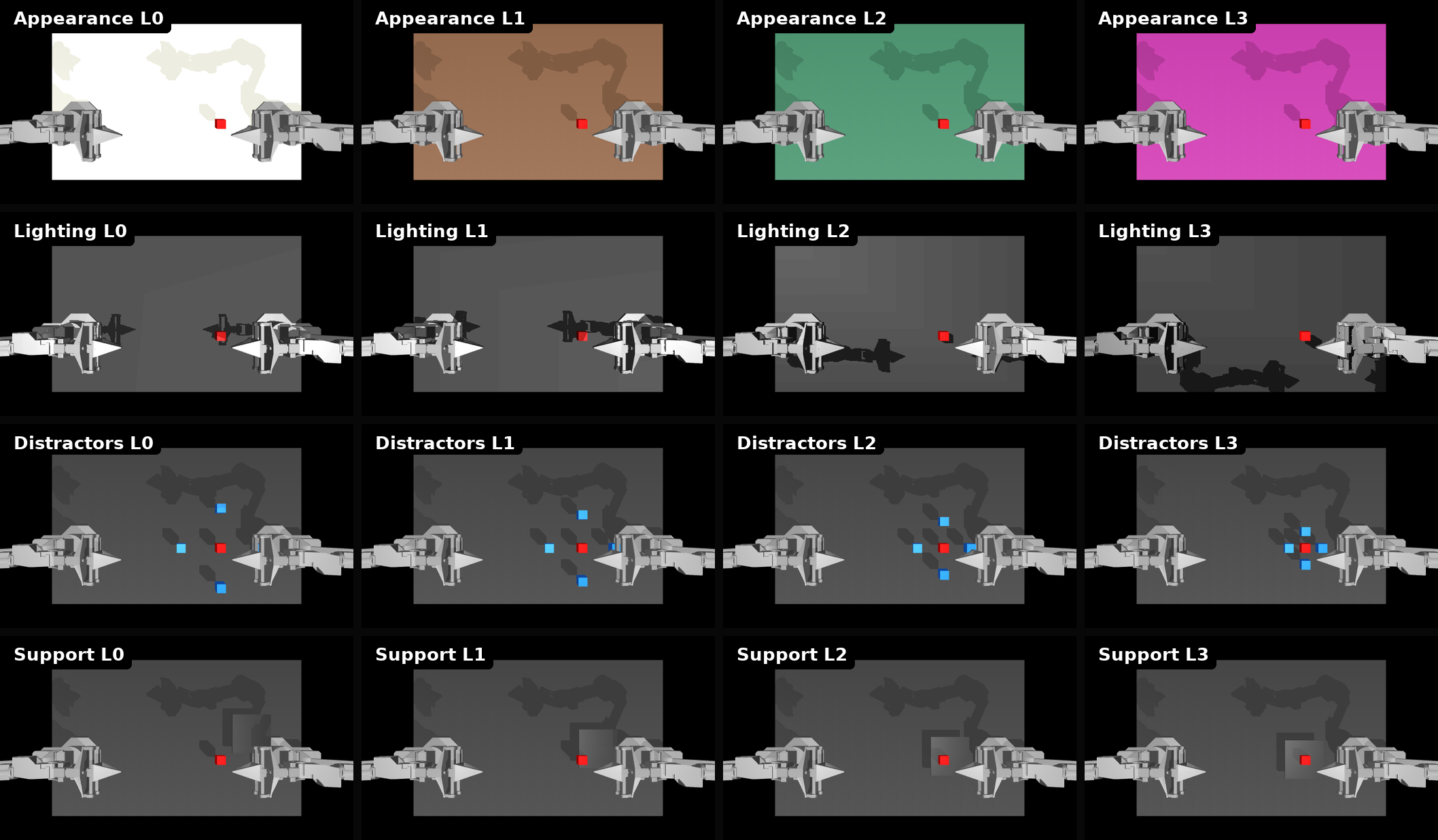}
    \caption{
    Single-factor nuisance grid for MuJoCo bimanual cube transfer. The same nuisance families are instantiated for the cube-transfer task to evaluate whether task-preserving visual shifts affect the nominal policy.
    }
    \label{fig:apx_transfer_cube_nuisance_grid}
\end{figure*}

\begin{figure*}[p]
    \centering
    \includegraphics[width=0.98\textwidth]{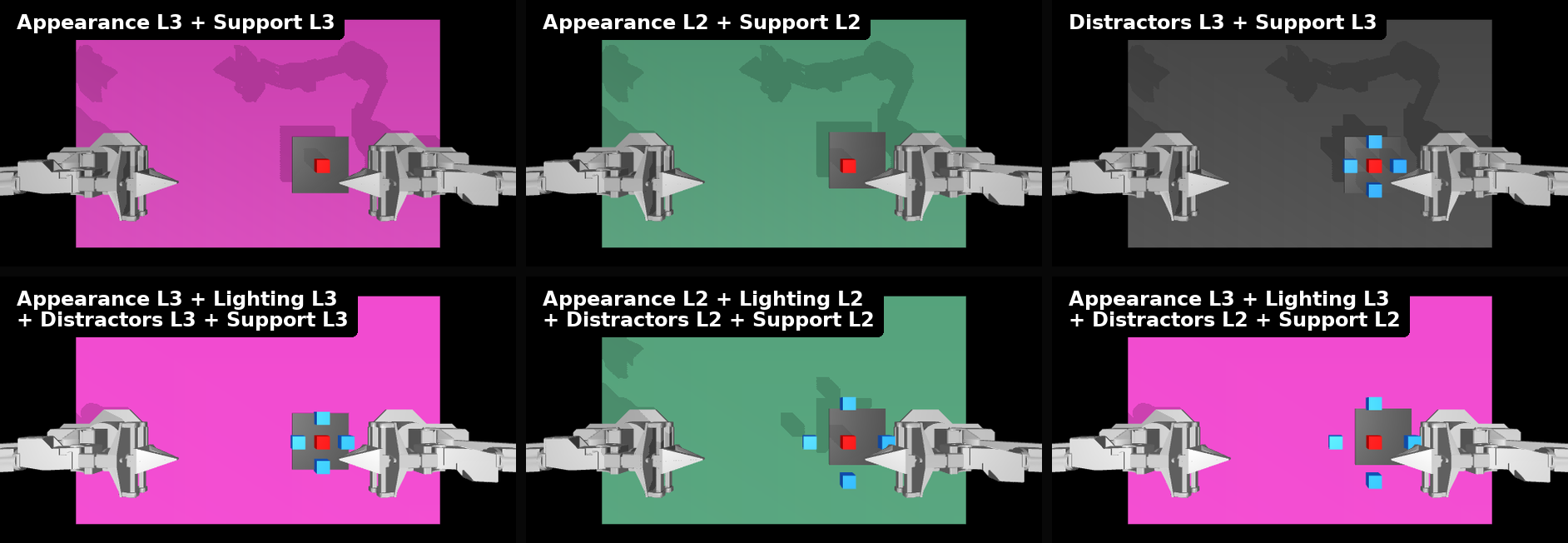}
    \caption{
    Multi-factor combination nuisance conditions for MuJoCo bimanual cube transfer. These predefined combinations test robustness under simultaneous task-preserving visual shifts.
    }
    \label{fig:apx_transfer_cube_combo_grid}
\end{figure*}

\begin{figure*}[p]
    \centering
    \includegraphics[width=0.9\textwidth]{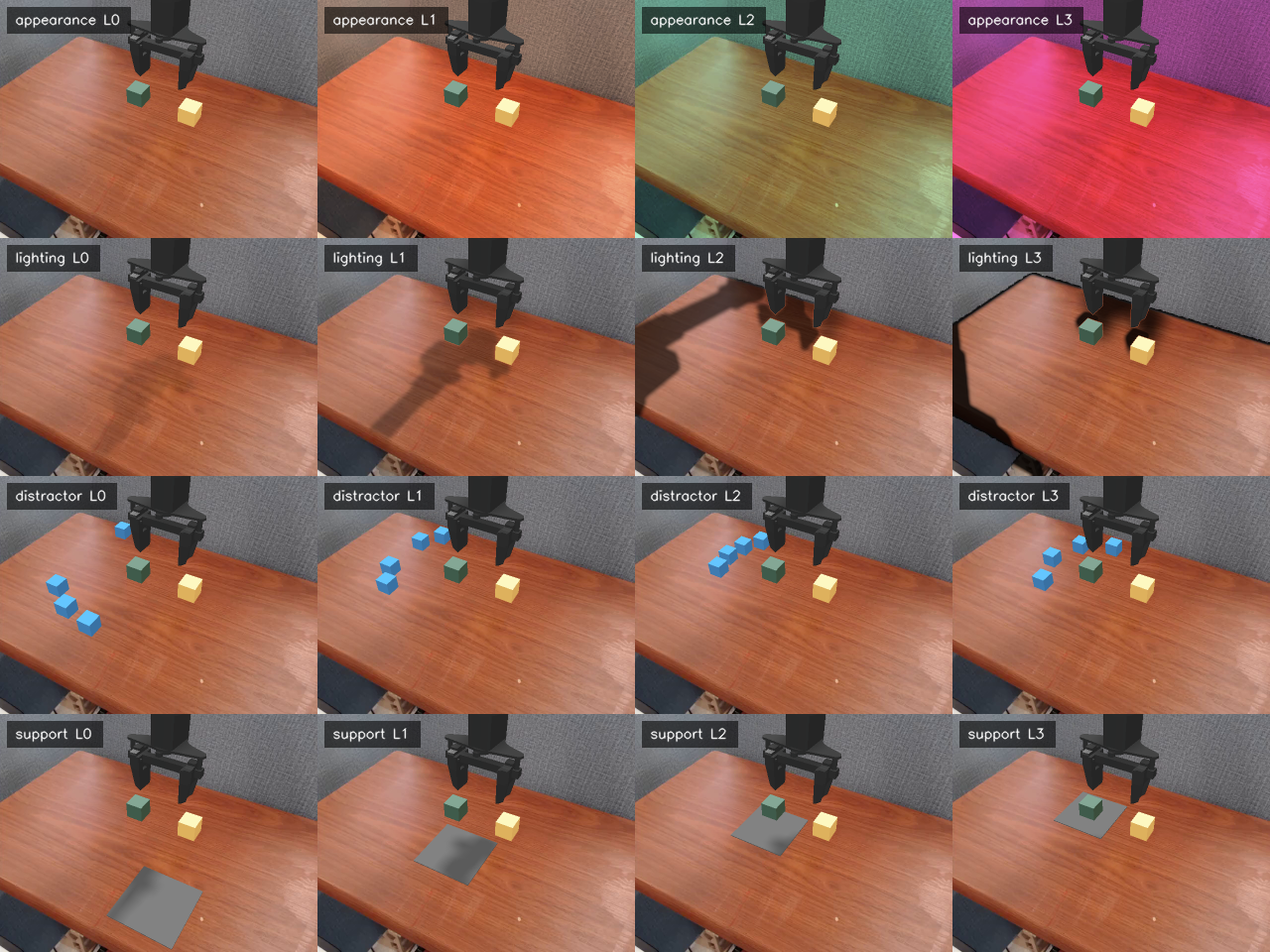}
    \caption{
    Single-factor nuisance grid for SimplerEnv cube stacking. The grid shows the appearance, lighting and shadow, distractor proximity, and local support nuisance families across their evaluated levels.
    }
    \label{fig:apx_stack_cube_nuisance_grid}
\end{figure*}

\begin{figure*}[p]
    \centering
    \includegraphics[width=0.9\textwidth]{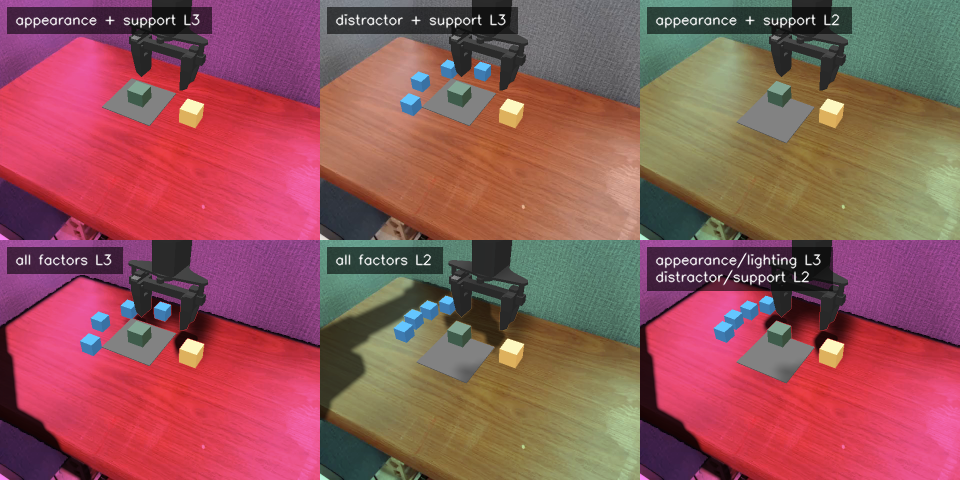}
    \caption{
    Multi-factor combination nuisance conditions for SimplerEnv cube stacking. These conditions combine multiple task-preserving visual perturbations and are used as predefined stress tests.
    }
    \label{fig:apx_stack_cube_combo_grid}
\end{figure*}

Figures~\ref{fig:apx_transfer_cube_nuisance_grid}--~\ref{fig:apx_stack_cube_combo_grid} show the full nuisance grids used in the experiments. The main paper shows a compact representative strip for readability, while the appendix provides the full visual set for both tasks.

\section{Narrow repair mainly repairs narrow failure modes}
\begin{figure*}[h]
    \centering
    \includegraphics[width=0.98\linewidth]{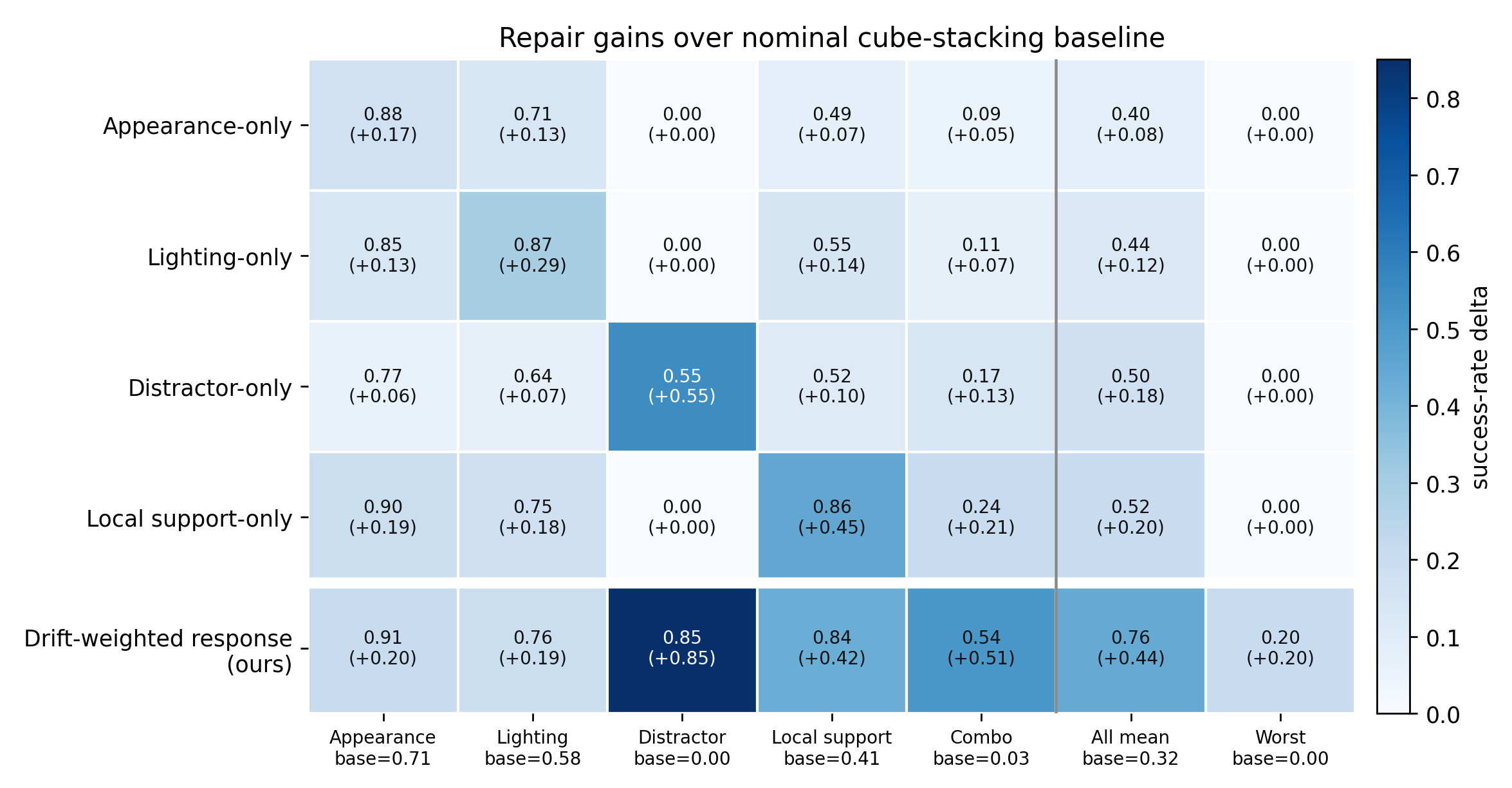}
    \caption{Narrow repair data repairs narrow failure modes. Cell colour shows the change in success rate relative to the nominal A200 SimplerEnv cube stacking policy, while cell text reports the final success rate with the change in parentheses. Each model receives the same budget of $K=30$ repair samples. The single-factor repair models use randomly selected demonstrations from one nuisance family, while the final row uses our proposed selection method with the same budget. Single-factor repair improves the targeted nuisance, but provides limited transfer to other factors. In contrast, our method achieves a broader repair over the mean and worst-case success.}
    \label{fig:single_factor_repair_heatmap}
\end{figure*}

A natural baseline for targeted repair is to add more demonstrations to the nuisance class where the model fails, for example, if the policy is weak to distractor objects, an operator will collect more demonstrations with distractors in the scene. While this strategy is intuitive, it assumes that any sample from the nuisance family is sufficient as a useful repair example. 
Figure~\ref{fig:single_factor_repair_heatmap} shows that single-factor repair can improve robustness to that class or incidentally related nuisances. However, it does not achieve broad robustness; this can be seen in that all single-factor repairs still have worst-case success at $0.0$, which indicates that the narrow repair leaves at least one nuisance condition unrepaired. In contrast, our drift-weighted response selection method achieves broader robustness, maintains comparable single-factor gains and raises worst-case success with the same repair budget.
These results suggest that semantic labels are useful for understanding failure, but they are not sufficient to select an efficient repair set on their own. Within a nuisance family, many examples may be redundant and adding multiple instances of the same failure mode gives little marginal benefit. By selecting demonstrations that cover diverse high-sensitivity conditions, our method can produce a more general repair rather than adding random instances from the fragile nuisance class.

\section{Held-Out Nuisance Generalisation} 
\label{Apx:HeldOut} 
\begin{figure*}[p] 
\centering 
    \includegraphics[width=0.98\textwidth]{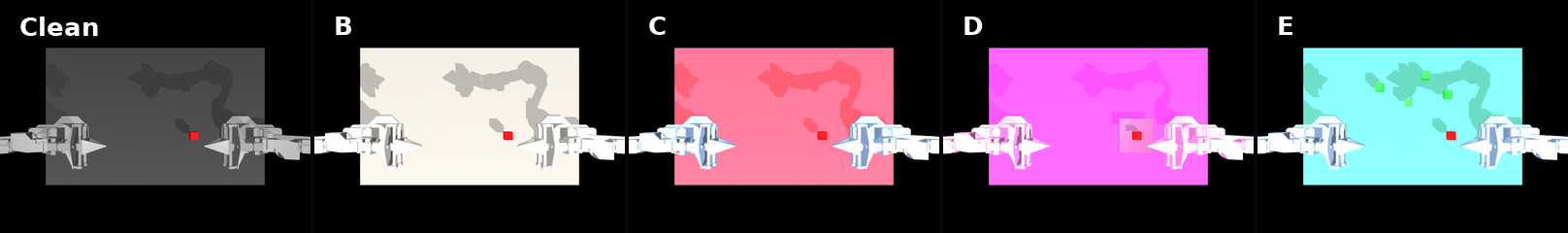} 
    \includegraphics[width=0.98\textwidth]{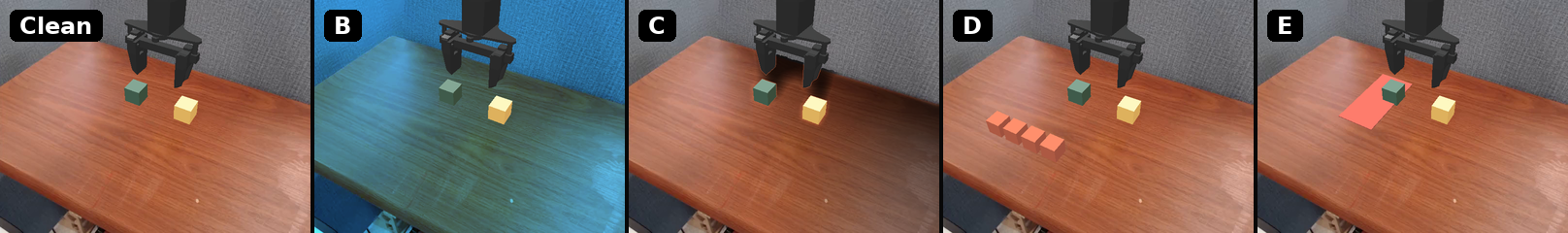} 
    \caption{ Representative held-out nuisance conditions for the two evaluation tasks. The top row shows MuJoCo cube-transfer held-out conditions, and the bottom row shows SimplerEnv cube-stacking held-out conditions. These visual nuisance instantiations are not used during candidate response generation, repair selection, or fine-tuning. } 
\label{fig:heldout_nuisance_examples} 
\end{figure*}

This appendix evaluates whether repair transfers to nuisance conditions not used during candidate response generation, repair selection, or fine-tuning. Held-out conditions use different visual instantiations of the same nuisance families, such as changed colours, lighting/shadow directions, distractor shapes, or distractor placements. These results test whether response-guided repair improves robustness beyond the exact nuisance values seen during repair. Figure~\ref{fig:heldout_nuisance_examples} shows illustrative examples of the heldout set with the initial results given in Table~\ref{tab:heldout_results}




\clearpage

\begin{table*}[p]
\centering
\caption{Held-out nuisance generalisation. Seen All nuis. reports performance on the 22 nuisance conditions used for the main evaluation. Held-out mean and Held-out worst report performance on nuisance instantiations not used during candidate response generation, repair selection, or fine-tuning. Gap is the difference between seen and held-out all-nuisance performance.}
\label{tab:heldout_results}
\small
\begin{tabular}{llrrrrr}
\toprule
Task & Method & \(K\) & Seen All nuis. & Held-out mean & Held-out worst & Held-out gap \\
\midrule
Cube transfer & Nominal & 0 & 0.30 & 0.02 & 0.02 & 0.28 \\
Cube transfer & Random & 20 & 0.56 & 0.11 & 0.03 & 0.44 \\
Cube transfer & Top-drift & 20 & 0.74 & 0.32 & 0.19 & 0.42 \\
Cube transfer & Response-guided (ours) & 20 & 0.96 & 0.57 & 0.43 & 0.38 \\
Cube transfer & Random (high-budget) & 500 & 1.00 & 0.90 & 0.80 & 0.10 \\
\midrule
Cube stacking & Nominal & 0 & 0.32 & 0.04 & 0.02 & 0.28 \\
Cube stacking & Random & 30 & 0.60 & 0.42 & 0.03 & 0.18 \\
Cube stacking & Top-drift & 30 & 0.63 & 0.44 & 0.00 & 0.19 \\
Cube stacking & Response-guided (ours) & 30 & 0.76 & 0.48 & 0.02 & 0.28 \\
Cube stacking & Random (high-budget) & 500 & 0.89 & 0.57 & 0.28 & 0.31 \\
\bottomrule
\end{tabular}
\end{table*}


\end{document}